\documentclass[a4paper,twoside]{article}

\usepackage{epsfig}
\usepackage{subcaption}
\usepackage{calc}
\usepackage{amssymb}
\usepackage{amstext}
\usepackage{amsmath}
\usepackage{amsthm}
\usepackage{multicol}
\usepackage{pslatex}
\usepackage{apalike}
\usepackage{algorithm2e}
\usepackage[bottom]{footmisc}
\usepackage{booktabs} 
\usepackage{hyperref}
\usepackage{tabularx}

\usepackage{SCITEPRESS}     

\begin{document}

\title{On Theoretically-Driven LLM Agents for \\ Multi-Dimensional Discourse Analysis}  

\author{\authorname{Maciej Uberna\sup{1}\orcidAuthor{0009-0006-8953-8270}, Michał Wawer\sup{1}\orcidAuthor{0009-0004-2717-1616} Jarosław A. Chudziak\sup{2}\orcidAuthor{0000-0003-4534-8652} and Marcin Koszowy\sup{1}\orcidAuthor{0000-0001-5553-7428}}
\affiliation{\sup{1}Laboratory of \textit{The New Ethos}, Warsaw University of Technology, Poland}
\affiliation{\sup{2}Faculty of Electronics and Information Technology, Warsaw University of Technology, Poland}
\email{\{maciej.uberna.dokt, michal.wawer.stud, jaroslaw.chudziak, marcin.koszowy\}@pw.edu.pl,}
}


\abstract{Identifying the strategic uses of reformulation in discourse remains a key challenge for computational argumentation. While LLMs can detect surface-level similarity, they often fail to capture the pragmatic functions of rephrasing, such as its role within rhetorical discourse. This paper presents a comparative multi-agent framework designed to quantify the benefits of incorporating explicit theoretical knowledge for this task. We utilise an dataset of annotated political debates to establish a new standard encompassing four distinct rephrase functions: Deintensification, Intensification, Specification, Generalisation, and Other, which covers all remaining types (D-I-S-G-O). We then evaluate two parallel LLM-based agent systems: one enhanced by argumentation theory via Retrieval-Augmented Generation (RAG), and an identical zero-shot baseline. The results reveal a clear performance gap: the RAG-enhanced agents substantially outperform the baseline across the board, with particularly strong advantages in detecting Intensification and Generalisation context, yielding an overall Macro F1-score improvement of nearly 30\%. Our findings provide evidence that theoretical grounding is not only beneficial but essential for advancing beyond mere paraphrase detection towards function-aware analysis of argumentative discourse. This comparative multi-agent architecture represents a step towards scalable, theoretically informed computational tools capable of identifying rhetorical strategies in contemporary discourse.}

\onecolumn \maketitle \normalsize \setcounter{footnote}{0} \vfill

\section{\uppercase{Introduction}}\label{sec:introduction}  

Argumentative discourse is characterised not only by the presentation of arguments \cite{Eemetal14} but also by the strategic use of rephrasing, i.e. the technique of altering a message for a rhetorical gain \cite{Younisetal2023}. Rephrasing constitutes a prevalent rhetorical and linguistic device employed to reinforce a message, tailor it to diverse audiences, or enhance its persuasive force. It may also serve to create an impression of argumentative richness by reiterating the same point in varied forms, thereby giving the appearance of multiplicity where there is, in fact, repetition. The capacity to reformulate utterances, while subtly altering their meaning plays a pivotal role in shaping discursive dynamics, influencing public perception, and structuring the flow of debate.

While theoretical frameworks for analyzing argumentation (e.g. based on Inference Aanchoring Theory \cite{Visseretal2018,LEWINSKI2013164}, argumentation schemes \cite{WalKra95}) and studies on rephrasing \cite{Kiljan2024,Budzynska2024} provide valuable insights, their application often remains manual or computationally limited. Existing argument mining techniques often struggle with the nuances of reformulation, potentially misclassifying strategic rephrases or failing to distinguish genuine support from manipulative restatements \cite{Konatetal2016}. This represents a significant gap in developing scalable computational tools for analysis of how rephrasing really functions in discourse.

To address this gap, we investigate whether grounding intelligent agents \cite{marreed2025enterprisereadycomputerusinggeneralist} in explicit argumentation and rhetorical theory significantly improves their ability to analyse how rephrasing is used, particularly in classification. Our key research question (RQ) is: Does the explicit knowledge of argumentation and rephrase theory lead to a demonstrable difference in an AI agent's analysis of reformulations, compared to agents relying solely on the implicit knowledge within pre-trained LLMs?

To answer this, we propose and implement a "Rephrasing Agents" platform. We simulate two architectures performing speech act analysis:
\begin{itemize}
    \item Theoretically-Informed Agents: Equipped via RAG with knowledge from argumentation theory and rephrase studies (including taxonomies and criteria for identifying misuses of rephrase \cite{Visseretal2018,Budzynska2024}). 

    \item Zero-Shot Agents: Identical LLM-based agents operating without access to the explicit theoretical knowledge base.
\end{itemize}

We test this system using transcribed US2016 Reddit and televised presidential debates based on \cite{visser2020argumentation} with 401 rephrase selected, a genre rich in persuasive and communicative manoeuvres where rephrasing plays a key role. This research direction, combined with creating annotated corpora of natural language discourse, e.g. \cite{visser2020argumentation} is a part of the `Argument Web' project for building an online ecosystem of devices and services for argumentation in the digital communication \cite{Reeetal17pat}. Among some recent findings in this area there are e.g. dialogical interfaces for exploring complex debates \cite{Lawrenceetal2022} or systems for detecting argumentative fallacies along with specifying challenges for the LLM-supported fallacy identification in the wild \cite{ruiz2023detecting}. This paper provides an empirically grounded understanding of how LLM-driven agents \cite{liu2025selfelicitlanguagemodelsecretly}, when enhanced with explicit theoretical knowledge, can simulate and analyse the complex dynamics of reformulation in argumentative discourse.

Integrating advanced NLP techniques within this theoretically-motivated MAS allows us to develop mechanisms for identifying how strategically rephrased statements potentially reinforce or mitigate polarising trends. Understanding and computationally exploring methods to identify indicators of misuse of reformulation represents an important step towards developing frameworks for next-generation AI tools capable of more effectively detecting forms of online harms \cite{guo2024investigationlargelanguagemodels}, such as subtle hate speech, disinformation campaigns, and the spread of fake news, which often rely on manipulative reframing rather than outright falsehoods. 

\section{\uppercase{Related Work}}\label{sec:RelatedWork}

Our approach integrates insights of rephrasing with a multi-agent system (MAS). This section provides the context for this integration by reviewing relevant prior work. 

\subsection{Rephrasing in Discourse}

\begin{figure*}[h]
    \centering
    \includegraphics[width=1\linewidth]{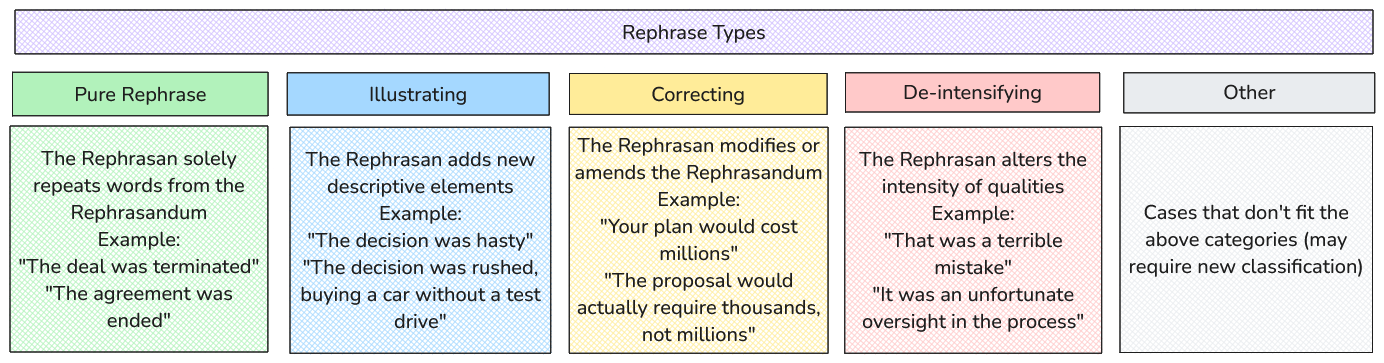}
    \caption{Classification of rephrase based on \cite{Kiljan2024}.}
    \label{fig:rephrase_classification}
\end{figure*}

"Rephrase in Argument Structure" \cite{Konatetal2016} is among the earliest studies to examine the role of rephrasing in argumentative discourse. The study highlights key challenges in argument mining, particularly in assessing the degree of support for a speaker’s position in dialogue. In this work, rephrasing is defined as two non-overlapping text spans that occupy the same position within the argument structure, thereby preserving its overall coherence (in other words, preserve the argument structure). Further examination of rephrasing \cite{Koszowyetal2022-rephrase} identifies linguistic patterns such as generalisation and specification across communicative genres. Their methodology combined annotation-based analysis to classify rephrase types and their locutionary and illocutionary aspects, with a participatory study exploring perlocutionary effects.

"Rephrasing is not arguing, but it is still persuasive" \cite{Younisetal2023} examines the rhetorical characteristics and persuasive impact of rephrasing in argumentative discourse through three experimental studies. The findings highlight the complexity of rephrasing, its rhetorical advantages in communication, and its potential to enhance both the persuasiveness of statements and the perceived trustworthiness of the speaker.  

\cite{Kiljan2024} introduces an annotation scheme for categorising rephrase types, shifting focus from static linguistic cues to the dynamic process within a single act of rephrasing. The authors propose five distinct categories based on two textual components: rephrasandum (the initial-input segment) and rephrasans (the reformulated-output segment), presented in Fig. \ref{fig:rephrase_classification}.

\cite{Budzynska2024} propose a model analysing rephrased arguments with emphasis on ethos-sentiment interaction across discourse contexts. They introduce \textit{DynRephAn}, a tool for capturing linguistic variations in ethos and sentiment during rephrasing, enabling automatic visualisation of these changes.

\subsection{Multi-Agent LLM-based Systems}

Applying intelligent agents to complex NLP tasks offers several advantages over monolithic approaches \cite{wang2025mixtureofagents,harbarChudziakOxfordStyleDebatesLLMs}, particularly when tackling pragmatic function identification. MAS involves designing systems composed of multiple autonomous agents that interact with each other and their environment to achieve individual or collective goals.

In the context of NLP \cite{Sadowski_2025}, this paradigm allows for a modular decomposition of intricate problems \cite{zhang2025agenticcontextengineeringevolving,chen2025theorymindlargelanguage}. Complex tasks, such as understanding argumentation or subtle speech act functions, can be broken down into sub-tasks, each handled by a dedicated agent \cite{kostkaChudziakCognitiveSynergyMAS}. For instance, one agent might focus solely on discourse segmentation, while others specialize in relation extraction, intent recognition, or accessing external knowledge. 

Furthermore, MAS facilitates the integration of diverse knowledge sources and reasoning mechanisms \cite{jimenezromero2025multiagentsystemspoweredlarge}. As demonstrated in our architecture, some agents can be equipped with specific knowledge (via RAG), while others operate based on general pre-trained knowledge, enabling comparisons and critiques within the system itself. This approach allows for building systems where agent interactions \cite{more_agents_2024}, potentially involving negotiation or critical evaluation (like our Critic), lead to more robust and explainable outcomes than a single, large model might produce \cite{Chung_2025}. The use of MAS-frameworks like CrewAI or AutoGen simplifies the orchestration of these interacting agents and the management of shared state, making the MAS approach practical for implementing theoretically-grounded NLP models.

\section{\uppercase{Hypothesis and Approach}}

The central computational problem addressed in this paper lies at the intersection of natural language understanding, argumentation, and rhetoric. While LLMs demonstrate capabilities \cite{chen-etal-2024-exploring-potential,wagner-ultes-2024-controllability}, their ability to discern the subtle pragmatic functions and potential misuses of reformulation solely from statistical patterns often falls short. Distinguishing legitimate clarification from manipulative distortion frequently requires an understanding of argumentative context in dialogue sequence and rhetorical principles potentially absent in standard zero-shot models \cite{wu2025largelanguagemodelssensitive,zhang2025achillesheeldistributedmultiagent}. This creates a gap: the lack of reliable computational methods for analysing how reformulation is used and identifying where it deviates from fair and reasoned discourse.

Based on this problem, we formulate the central hypothesis of this research, focusing on the contribution of our complete multi-agent architecture: \textit{A MAS whose specialist agents are explicitly grounded in argumentation theory via RAG will achieve a demonstrably higher classification accuracy—particularly rephrase categories—than an identical system operating in a zero-shot capacity.}
We hypothesise that the informed system's advantage will stem from its ability to apply theoretical principles, whereas the zero-shot system will be limited to surface-level linguistic patterns.

\section{\uppercase{Methodology}}\label{sec:SystemArchitecture} 

This section introduces the Rephrasing Agents platform, showcasing the multi-agent architecture developed for this research. We will outline the workflow, agent interactions, and state management that enable the comparative reformulation analysis.

\subsection{Rephrase Classification} 
\label{sec:rephrase_theory}

\begin{table*}[t]
    \caption{Rephrase categories with definitions and examples from political discourse.}
    \centering
    \label{tab:rephrase_categories}
    \small
    \begin{tabularx}{\textwidth}{l p{4cm} X}
    \toprule
    \textbf{Category} & \textbf{Definition} & \textbf{Example} \\
    \midrule
    De-intensification & Weakens a point & 
    \textit{``Anderson Cooper is really doing a great job''} $\rightarrow$ \textit{``Anderson Cooper is not bad at all''} \\
    \addlinespace
    Intensification & Strengthens a point by reinforcing qualities & 
    \textit{``Anderson Cooper is really doing a great job''} $\rightarrow$ \textit{``Anderson Cooper is killing it''} \\
    \addlinespace
    Specification & Adds detail or narrows scope & 
    \textit{``there is nothing to be impressed with about Hillary''} $\rightarrow$ \textit{``nothing impressive [...] other than her ambition, and her pandering''} \\
    \addlinespace
    Generalisation & Broadens or abstracts the original point & 
    \textit{``TRUMP is not answering the question''} $\rightarrow$ \textit{``TRUMP never did answer the question''} \\
    \addlinespace
    Other & Rephrase types not covered & 
    \textit{``cutting mics should totally be a thing''} $\rightarrow$ \textit{``I wish the organisers would make cutting microphones part of the debate''} \\
    \addlinespace
    Not a Rephrase & Pairs that do not constitute reformulation (e.g., inferences) & 
    \textit{``Tariffs will cause companies to reopen plants''} $\rightarrow$ \textit{``we have to stop them from leaving''} \\
    \bottomrule
    \end{tabularx}
\end{table*}

We propose a two-step method to classify and evaluate types of rephrasing. Firstly, rephrases are categorised by the MAS system into one of five D-I-S-G-O types or identified as not a rephrase. Each category is defined by specific rules guiding the LLM. MAS examines the input and output phrases row-by-row, alongside their ilocutions (i.e. reconstructed meanings by an annotator), to assign the appropriate category.

Secondly, when a reformulation is detected and classified, the rephrase is compared to a gold standard annotated from the US 2016 presidential debates corpus. A total of 460 rephrase instances were manually classified using this scheme, with each instance assigned a single rephrase category. This corpus serves as input for the experiments detailed in Section \ref{sec:experimental_setup}.

\subsection{Multi-Agent Architecture}

\begin{figure*}[ht]
    \centering
        \includegraphics[width=0.85\linewidth]{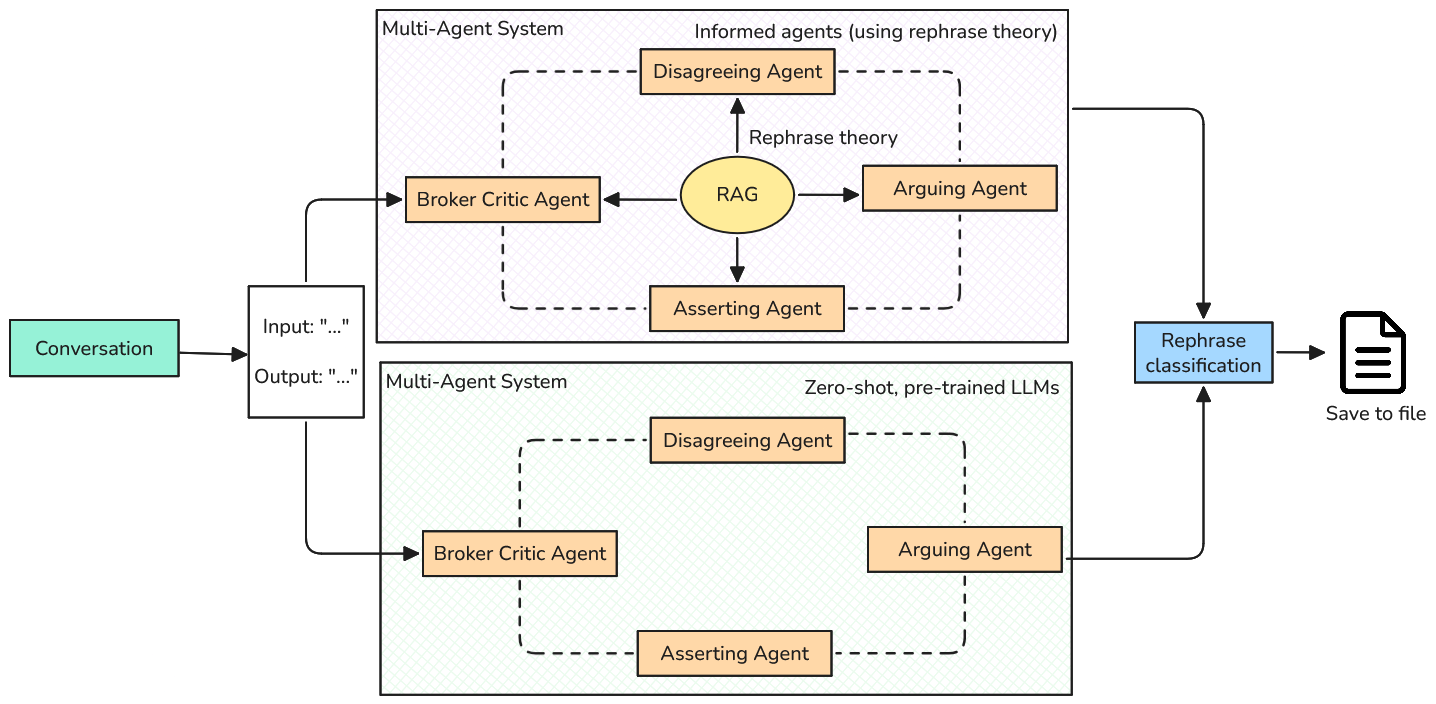}
    \caption{MAS architecture. The Informed System (top) equips agents with RAG access to argumentation theory, while the Zero-Shot System (bottom) relies solely on pre-trained LLM knowledge. Both share identical agent roles and orchestration.}
    \label{fig:system_architecture}
\end{figure*}

The core of our methodology is a parallel evaluation of two distinct MAS: an \textit{Informed System} and a \textit{Zero-Shot System}. As illustrated in Figure~\ref{fig:system_architecture}, both systems share an identical architecture but differ in their access to theoretical knowledge. Each system is composed of four specialised agents \cite{react_agent} designed to collaborate on the analysis of a given rephrase pair (a rephrasandum and a rephrasans).

All prompts used in our paper are available in github repository \footnote{Code available at \href{https://github.com/michal-wawer/RephraseAnalysisAgents_ICAART2026}{github.com/michal-wawer/RephraseAnalysisAgents\_ICAART2026}}. The agent team within each system includes:
\begin{itemize}
    \item \textbf{Asserting Agent:} Responsible for making factual, evidence-based assertions about the linguistic and semantic relationship between the two statements.
    \item \textbf{Arguing Agent:} Tasked with constructing a logical argument for a specific classification, drawing inferences and building a coherent case.
    \item \textbf{Disagreeing Agent:} Designed to critically evaluate the ongoing analysis, challenge assumptions, and identify potential misuses or hidden complexities that other agents might overlook.
    \item \textbf{Broker Critic Agent:} Serves as the orchestrator and final arbiter of the group. It manages the conversational flow, synthesizes the perspectives of the other agents, and produces the definitive classification and a summary justification.
\end{itemize}

The analytical process is structured as a moderated conversation. The Broker Critic Agent initiates and guides the discussion, calling upon the specialist agents to contribute their analysis in sequence. After a predetermined number of conversational turns, the Broker synthesizes the discussion and provides a final classification. 

\subsection{System Implementation}

We implemented our experimental setup using Python based CrewAI framework, which provides capabilities for orchestrating conversations between multiple LLM-based agents. The Broker Critic Agent was configured as manager agent, which is responsible for directing the conversational flow and terminating the interaction upon reaching a conclusion. The underlying LLM for all agents was OpenAI's ``GPT-5 Mini``. This model was selected for its favorable balance of performance and efficiency. It offers reasoning capabilities comparable to top-tier LLMs but at a significantly higher speed and lower cost.

\subsubsection{Retrieval-Augmented Generation Configuration}
The key distinction between our two systems is implementation of the RAG component, which was exclusively provided to the agents of the Informed System. The RAG knowledge base was constructed from a curated set of academic texts on argumentation theory and rephrase analysis. The contents included:

\begin{itemize}
    \item \textbf{Theoretical Frameworks:} Core principles from Inference Anchoring Theory (IAT), including the structural characterization of rephrase relations \cite{Visseretal2018,Koszowyetal2022-rephrase}.
    \item \textbf{Rephrase Taxonomies:} Classifications of different rephrase types (e.g., intensification, specification) and their linguistic markers \cite{Younisetal2023,Kiljan2024,Konatetal2016}, based on the annotation scheme underpinned in section \ref{sec:rephrase_theory}.
\end{itemize}

\subsection{Comparative Analysis Approach}

Our experimental design is centered on a direct comparison between the outputs of two parallel MAS. The systems operate in complete isolation for each rephrase pair, neither system has access to the process or results of the other. This strict separation ensures that we can isolate and measure the impact of the single variable under investigation: access to explicit theoretical knowledge.

\begin{itemize}
    \item \textbf{The Informed System:} In this configuration, the Asserting, Arguing, and Disagreeing agents were equipped with the RAG tool. Their system prompts explicitly instructed them to ground their analysis in the provided theoretical knowledge base, referencing it to support their claims.

    \item \textbf{The Zero-Shot System:} This system served as our baseline. Its agents were identical to their Informed counterparts in every respect—including their prompts and the underlying LLM—but they were not given access to the RAG tool. Their analysis, therefore, relied solely on the implicit knowledge contained within the pre-trained model's parameters.
\end{itemize}

\section{\uppercase{Experimental Setup}}
\label{sec:experimental_setup}

To validate our central hypothesis, we conducted a set of experiments designed to quantitatively measure and qualitatively analyze the impact of theoretical grounding on a MAS ability to classify rephrase functions. This section details our experimental setup, the dataset used, and the evaluation metrics.

\subsection{Dataset}

The evaluation is based on a gold-standard dataset derived from the US2016 corpus \cite{visser2020argumentation}, which comprises professionally annotated rephrase relations from the 2016 U.S. presidential debates. A team of two annotators extracted 464 rephrase pairs from this corpus (input and output) and independently classified each rephrase into one of six categories. Following this, Cohen’s kappa coefficient was calculated to assess inter-annotator agreement, as presented in Table~\ref{tab:kappatable}. For our automated classification experiments, we selected 401 of these annotated pairs.

\begin{table}[h!]
    \centering
    \caption{Inter-annotator agreement for rephrase classification.}
    \label{tab:kappatable}
    \resizebox{0.85\columnwidth}{!}{
        \begin{tabular}{@{}lccc@{}}
        \toprule
        \textbf{Category} & \textbf{IAA Cohen's $\kappa$} \\ \midrule
        \quad Deintensifying & 44\% \\
        \quad Intensifying & 37\% \\
        \quad Specification & 48\% \\
        \quad Generalising & 39\% \\
        \quad Other & 17\% \\
        \quad No\_rephrase & 31\% \\ \midrule
        \quad \textbf{Overall} & \textbf{36\%} \\ \midrule
        \end{tabular}
    }
\end{table}

\subsection{Baselines and Comparative Systems}

Our experimental design is centred on a direct comparison between two MAS to isolate the impact of theoretical grounding. To disentangle the individual contributions of RAG enhancement and multi-agent collaboration, we conduct a 2×2 study. The four configurations vary along two dimensions: agent count (single vs. multi-agent) and knowledge access (zero-shot vs. RAG-enhanced). Single-agent configurations use the Broker Critic Agent in isolation, while multi-agent configurations employ the full four-agent architecture. RAG-enhanced variants have access to the curated theoretical knowledge base, whereas zero-shot variants rely solely on the pre-trained LLM. This design isolates whether performance gains stem primarily from theoretical knowledge, multi-agent deliberation, or their combination.

\subsection{Evaluation Metrics}
\label{sec:metrics}

For the seven-category classification task, we report the Macro Average F1-Score, which calculates the metric independently for each class and then takes the unweighted average. This treats all classes as equally important, regardless of their frequency (support). Additionally, to provide a single measure that is particularly well-suited for potentially imbalanced multi-class problems, we also calculate the Matthews Correlation Coefficient (MCC). It takes into account true and false positives and negatives for all classes simultaneously, producing a balanced measure even when class sizes differ substantially (as seen in our dataset, where 'Specification' has 137 instances while 'Other' has only 35). An MCC score of +1 indicates perfect agreement, 0 indicates performance no better than random, and -1 indicates total disagreement.

\section{\uppercase{Results}}

The experimental results demonstrate clear performance differences across all four configurations, supporting our hypothesis that theoretical grounding is critical for rephrase analysis.

Table~\ref{tab:main_performance_with_mcc} presents the comparative F1-scores for the six-category rephrase classification task. The RAG-enhanced MAS achieved the highest performance, with a macro-averaged F1-score of 0.67 and MCC of 0.64, indicating strong correlation with ground truth. In contrast, the Single Zero-Shot baseline achieved only 0.27 F1 and 0.16 MCC, performing only marginally better than random classification.
\begin{table}[h]
    \centering
    \caption{Comparative results across four configurations.}
    \label{tab:main_performance_with_mcc}
    \setlength{\tabcolsep}{4pt}
    \resizebox{\columnwidth}{!}{
    \begin{tabular}{l c c c c}
    \toprule
    & \multicolumn{2}{c}{\textbf{Single Agent}} & \multicolumn{2}{c}{\textbf{MAS}} \\
    \cmidrule(lr){2-3} \cmidrule(lr){4-5}
    \textbf{Category} & 0-Shot & RAG & 0-Shot & RAG \\
    \midrule
    Deintensifying & 0.26 & 0.55 & 0.33 & 0.68 \\
    Intensifying   & 0.24 & 0.61 & 0.31 & 0.77 \\
    Specification  & 0.40 & 0.63 & 0.48 & 0.74 \\
    Generalising   & 0.33 & 0.68 & 0.42 & 0.81 \\
    Other          & 0.11 & 0.21 & 0.18 & 0.30 \\
    No rephrase    & 0.28 & 0.59 & 0.35 & 0.73 \\
    \midrule
    \textbf{Macro F1} & 0.27 & 0.54 & 0.38 & 0.67 \\
    \textbf{MCC} & 0.16 & 0.48 & 0.24 & 0.64 \\
    \bottomrule
    \end{tabular}
    }
\end{table}
The 2×2 design isolates the contributions of each component. Adding RAG to a single agent yields a improvement of +0.27 F1, demonstrating that theoretical grounding alone provides significant benefit. Adding multi-agent collaboration without RAG contributes a more modest +0.11 F1. Notably, the full MAS+RAG configuration achieves +0.40 F1 over the baseline, exceeding the sum of individual contributions (+0.38), suggesting that RAG and multi-agent deliberation interact synergistically—the MAS architecture amplifies the benefits of theoretical knowledge by enabling specialist agents to apply it from multiple analytical perspectives.

\begin{figure}[ht]
    \centering
        \includegraphics[width=0.85\linewidth]{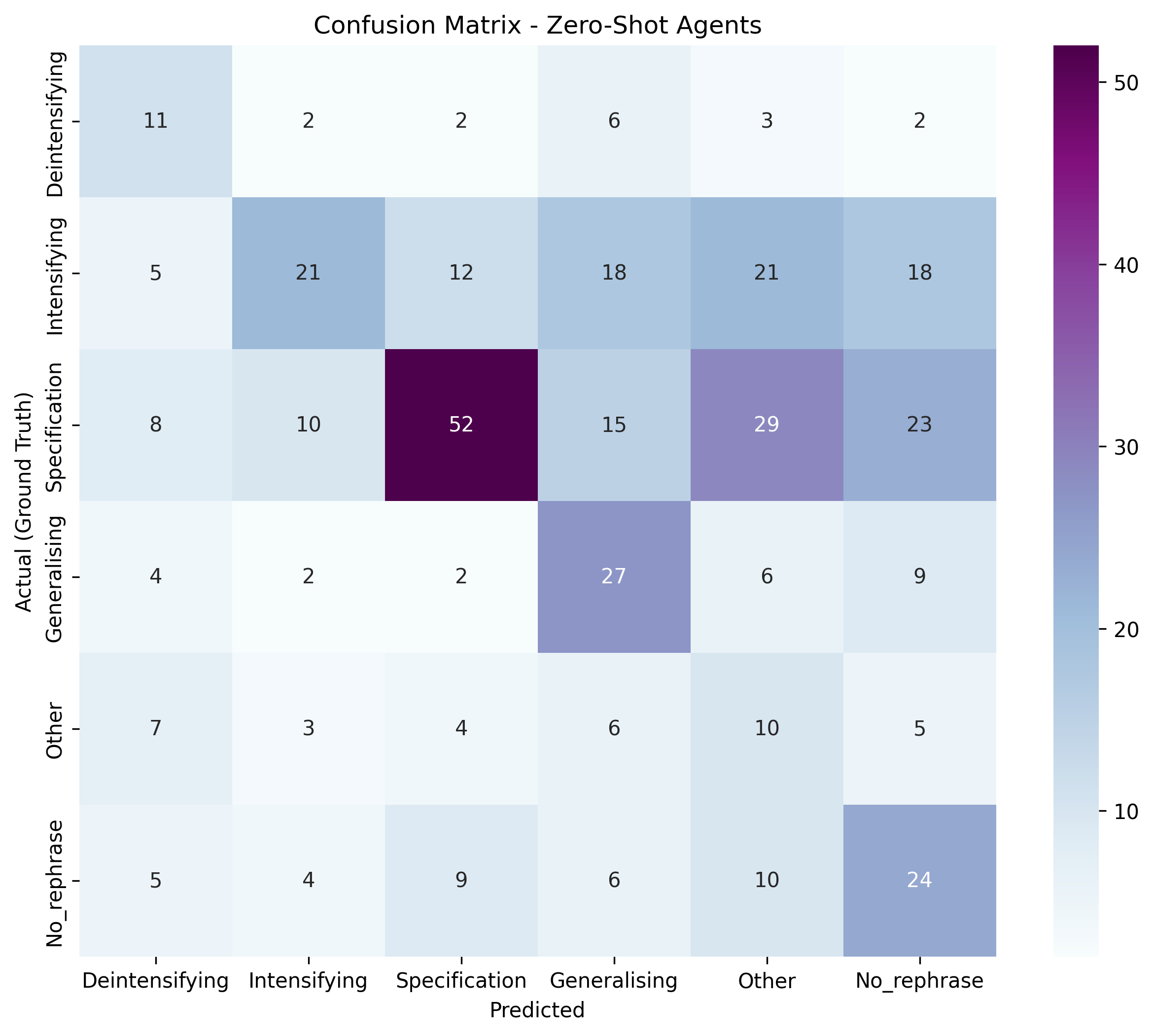}
    \caption{Confusion matrix of zero-shot MAS.}
    \label{fig:matrix_zero_shot}
\end{figure}

\begin{figure}[ht]
    \centering
        \includegraphics[width=0.85\linewidth]{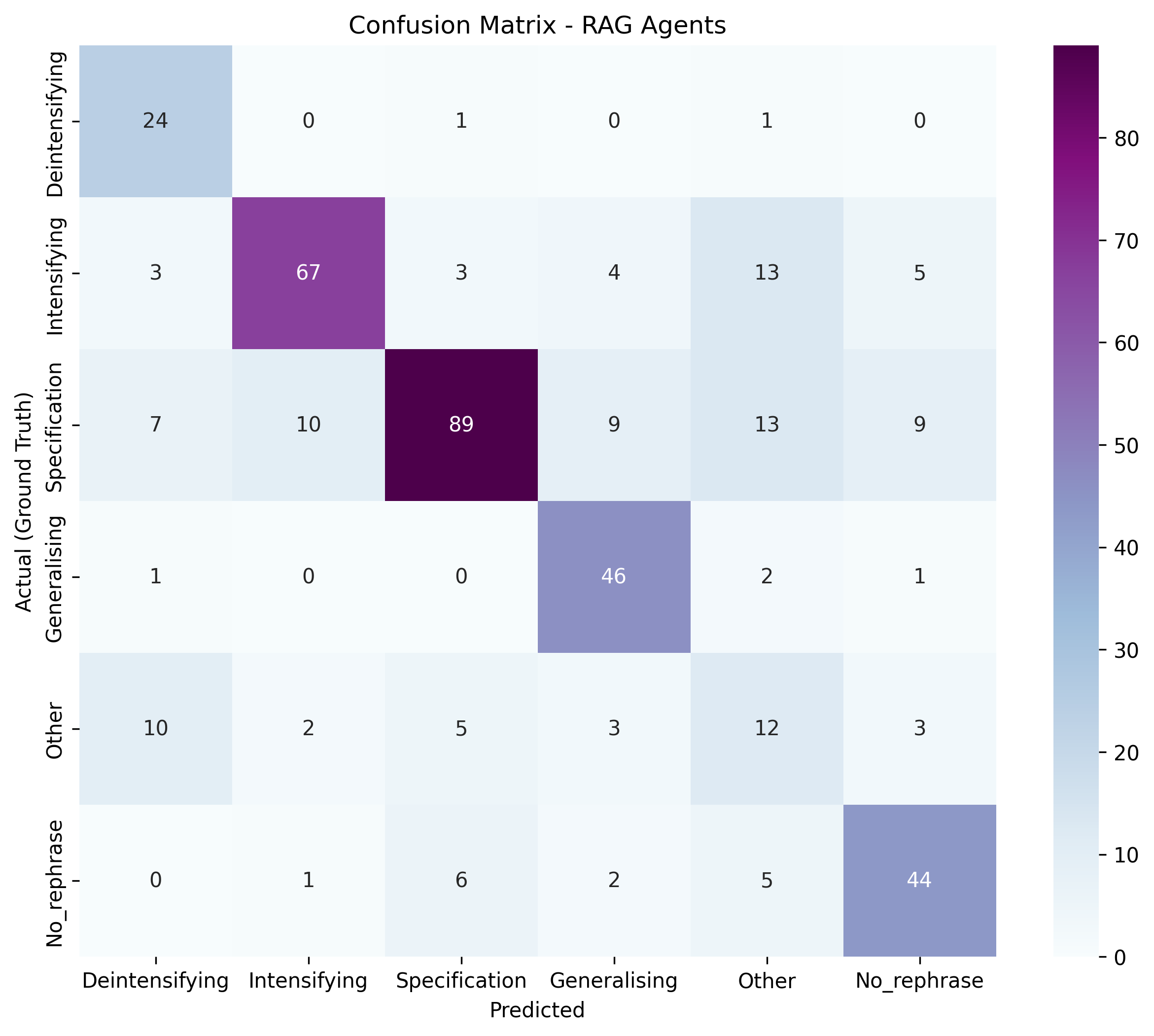}
    \caption{Confusion matrix of RAG enhanced MAS.}
    \label{fig:matrix_rag}
\end{figure}

Performance gains were most pronounced for categories requiring deeper pragmatic understanding. In the `Generalising' category, MAS+RAG scored 0.81 compared to 0.33 for Single Zero-Shot—an improvement of nearly 150\%. The corresponding confusion matrices in Figure~\ref{fig:matrix_zero_shot} and ~\ref{fig:matrix_rag} illustrate this disparity. RAG-enhanced predictions concentrate along the diagonal, while Zero-Shot predictions show widespread confusion between functionally distinct categories.

\section{\uppercase{Discussion and Future Work}}\label{sec:discussion}

Our RQ asked to what extent grounding a MAS in argumentation theory via RAG improves its performance compared to a zero-shot baseline. The RAG-Enhanced system achieved a macro-averaged F1-score of 0.670, nearly doubling the baseline's score of 0.375. As stated in introduction, a key challenge in computational argumentation is moving beyond surface-level similarity to capture the pragmatic function of a rephrase. Our findings demonstrate that the Zero-Shot system, while capable of some classification, struggled precisely with this task. As evidenced by its low F1-scores in categories such as `Intensifying` and `No\_rephrase`. A closer examination of misclassified rephrases reveals that the LLM confuses Deintensification with Intensification, as illustrated in the following example:\\
\\
\textbf{input:} \textit{Hillary Clinton is kicking Bernie Sanders' ass}\\
\textbf{output:} \textit{in this debate though Clinton is winning despite Bernie Sanders} \\

This rephrasing was annotated as Deintensification by human annotators but classified as Intensification by the MAS system. This discrepancy suggests that LLMs require additional fine-tuning, particularly with respect to specific idioms and phrases. 

This directly supports our central hypothesis: the Informed system's advantage stems from its ability to apply theoretical principles, whereas the Zero-Shot system is limited to surface-level linguistic patterns. The confusion matrices (Figure \ref{fig:matrix_zero_shot} and \ref{fig:matrix_rag}) visually corroborate this. The RAG-Enhanced system's predictions are tightly clustered along the diagonal, indicating correct classifications grounded in the provided rephrase taxonomies. In contrast, the Zero-Shot predictions are diffuse, showing widespread confusion between functionally distinct categories such as `Intensifying` and `Specification`. While both systems found the `Other` category challenging, the overall results confirm that theoretical grounding is not just beneficial but necessary for building function-aware analytical tools, thus addressing the gap identified in the introduction. 

While this study successfully demonstrates the value of a theoretically-grounded approach, it also opens several avenues for future research. Our main focus it to expand the theoretical scope, refine the agentic architecture, and extend the evaluation to more complex fallacies.
A key direction for our future work is to address the detection of manipulative rephrase, such as the straw man fallacy. Our current research efforts are focused on the manual creation of a high-quality, annotated gold-standard corpus specifically for this task. Once this dataset is complete, we will adapt the multi-agent framework presented here to automate the identification of these complex fallacies at scale. Ultimately, our goal is to apply rephrase-aware LLM agents to misinformation detection, where strategic reformulation serves as a primary vector for spreading false narratives online.

\section{\uppercase{Conclusion}}\label{sec:conclusion}

This paper was motivated by the gap between the theoretical understanding of reformulation and its computational analysis. As outlined in the introduction, existing methods often fail to distinguish legitimate clarification from manipulative distortion \cite{wagner-ultes-2024-controllability}. Our research sought to resolve this by investigating whether explicit theoretical knowledge could enhance the analytical capabilities of LLM-based agents. The experimental results confirm this unequivocally; the RAG-Enhanced system significantly outperformed the baseline across nearly all rephrase categories, demonstrating a clear advantage derived from its access to argumentation theory.

This study's central contribution is the empirical demonstration that theoretical grounding enables a qualitative shift in analytical capability—from surface-level pattern matching to deeper, function-aware reasoning. By presenting a comparative multi-agent framework and validating it on a human annotated dataset, we offer both a methodology and concrete evidence for the necessity of integrating domain-specific knowledge into LLM-based systems for discourse analysis. This work create grounding for more reliable and interpretable AI tools that can accurately identify the strategic and rhetorical uses of rephrasing in contemporary communication.

\section*{\uppercase{Acknowledgements}}

The work reported in this paper was supported by the Polish National Science Centre under grant 2020/39/I/HS1/02861.

\bibliographystyle{apalike}
{\small
\bibliography{./references}}

\end{document}